\relax
\documentclass[letterpaper]{article}
\usepackage{stylefile}
\usepackage{times}
\usepackage{helvet}
\usepackage{courier}
\usepackage[hyphens]{url}
\usepackage{graphicx}
\usepackage{adjustbox}
\usepackage{xcolor}
\usepackage{xspace}
\usepackage{amsmath}
\usepackage{appendix}
\usepackage{multirow}
\usepackage{color, colortbl}

\usepackage{graphicx}
\usepackage{natbib}
\usepackage{caption}
\newcommand{\MethodName}{BENN\xspace}
\newcommand{\etal}{\emph{et al.} }
\definecolor{Gray}{gray}{0.9}

\begin{document}
\title{\MethodName: Bias Estimation Using Deep Neural Network}

\author
{Amit Giloni,\thanks{Corresponding author}\textsuperscript{\rm 1}
Edita Grolman,\textsuperscript{\rm 1}
Tanja Hagemann,\textsuperscript{\rm 2,3}
Ronald Fromm,\textsuperscript{\rm 3}
Sebastian Fischer,\textsuperscript{\rm 2,3} \\
Yuval Elovici,\textsuperscript{\rm 1}
Asaf Shabtai\textsuperscript{\rm 1}\\
}

\affiliations {
    \textsuperscript{\rm 1} Department of Software and Information Systems Engineering, Ben-Gurion University of the Negev \\
    \textsuperscript{\rm 2} Technische Universitat Berlin \\
    \textsuperscript{\rm 3} Deutsche Telekom AG \\
    \{hacmona, edita\}@post.bgu.ac.il, tanja.hagemann@tu-berlin.de,\{ronald.fromm, sebastian-fischer\}@telekom.de, \\
    \{elovici, shabtaia\}@bgu.ac.il
}

\maketitle

\begin{abstract}
\begin{quote}
The need to detect bias in machine learning (ML) models has led to the development of multiple bias detection methods, yet utilizing them is challenging since each method: 
\textit{i)} explores a different ethical aspect of bias, which may result in contradictory output among the different methods, 
\textit{ii)} provides an output of a different range/scale and therefore, can't be compared with other methods,
and \textit{iii)} requires different input, and therefore a human expert needs to be involved to adjust each method according to the examined model.
In this paper, we present \MethodName -- a novel bias estimation method that uses a pretrained unsupervised deep neural network.
Given a ML model and data samples, \MethodName provides a bias estimation for every feature based on the model's predictions. 
We evaluated \MethodName using three benchmark datasets and one proprietary churn prediction model used by a European Telco and compared it with an ensemble of 21 existing bias estimation methods.
Evaluation results highlight the significant advantages of \MethodName over the ensemble, as it is generic (i.e., can be applied to any ML model) and there is no need for a domain expert, yet it provides bias estimations that are aligned with those of the ensemble.
\end{quote}
\end{abstract}

\section{Introduction}
Many new and existing solutions and services use machine learning (ML) algorithms for various tasks.
Induced ML models are prone to learning real-world behavior and patterns, including unethical discrimination and though inherit bias.
Unethical discrimination may have legal implications~\cite{malgieri2020concept}; for example, the European General Data Protection Regulation (GDPR) states that the result of personal data processing should be fair; consequently, the output of the induced ML model should not present any unethical bias.
Yet, underlying bias exists in various domains, such as facial recognition~\cite{buolamwini2018gender}, object detection~\cite{wilson2019predictive}, commercial advertisements~\cite{ali2019discrimination}, healthcare~\cite{obermeyer2019dissecting}, recidivism prediction~\cite{chouldechova2017fair}, and credit scoring~\cite{li2017reject}.

In order to detect this underlying bias, various methods have been proposed for bias detection and estimation~\cite{hardt2016equality,feldman2015certifying,berk2018fairness,verma2018fairness,narayanan2018translation,chouldechova2017fair}.
However, these approaches are not applicable to real life settings for the following reasons:
\textit{i)} Most methods produce binary output (bias exists or not); therefore, comparing the level of bias detected in different models and features is not feasible. 
\textit{ii)} While there are many bias detection and estimation methods, each explores a different ethical aspect of bias, which may result in contradictory output among the different methods, i.e., one method might determine that the examined ML model is fair, and another might detect underlying bias.
Therefore, in order to ensure that bias is not present in an induced ML model, the best practice is to apply an ensemble of all methods.
\textit{iii)} Applying an ensemble of all methods is a challenging task, since the methods need to be scaled to produce consistent bias estimations (using the same scale and range).
\textit{iv)} Different methods may require different data parameters as input. 
This necessitates a domain expert to determine which methods can be applied to the examined ML model, task, data, and use case, i.e., manual and resource consuming analysis.
For example, a method which uses the ground truth labels of samples cannot be used to evaluate an unsupervised ML model.

In this paper, we present BENN, a novel method for bias estimation that uses an unsupervised deep neural network (DNN).
Given an ML model and data samples, BENN performs a comprehensive bias analysis and produces a single bias estimation for each feature examined.
BENN is composed of two main components. 
The first component is a bias vector generator, which is an unsupervised DNN with a custom loss function.
Its input is a feature vector (i.e., a sample), and its output is a bias vector, which indicates the degree of bias for each feature according to the input sample. 
The second component is the post-processor, which, given a set of bias vectors (generated by the bias vector generator), processes the vectors and provides a final bias estimation for each feature.

Note that all bias detection and estimation methods are based on the ``fairness through unawareness" principle~\cite{verma2018fairness}, which means that changes in feature with ethical significance should not change the ML model's outcome.
 
Existing methods examine only one ethical aspect of this principle, whereas BENN evaluates all ethical aspects by examining how each feature affects the ML outcomes. 

We empirically evaluated BENN on three bias benchmark datasets: the ProPublica COMPAS~\cite{angwin2016machine}, Adult Census Income~\cite{blake1998adult}, and Statlog (German Credit Data)~\cite{kamiran2009classifying} datasets.
In addition, we evaluated BENN on a proprietary churn prediction model used by a European Telco, and used synthetic dataset that includes a biased feature and a fair one, allowing us to examine BENN in extreme scenarios.
The results of our evaluation indicate that BENN's bias estimations are capable of revealing model bias, while demonstrating similar behavior to existing methods. 
The results also highlight the significant advantages of BENN over existing methods; these advantages include the fact that BENN is generic and its application does not require a domain expert.
Furthermore, BENN demonstrated similar behavior to existing methods after applying a re-weighting mitigation method on the models and datasets to reduce the unwanted bias.

The main contributions of this paper are as follows:
\begin{itemize}
    \item To the best of our knowledge, BENN is the first bias estimation method which utilizes an unsupervised deep neural network.
    Since DNNs are able to learn significant patterns within the data during training, BENN performs a more in depth bias examination than existing methods.
    \item In contrast to all other methods which focus on just one ethical aspect, BENN performs a comprehensive bias estimation based on all of the ethical aspects currently addressed in the literature.
    \item BENN is a generic method which can be applied to any ML model, task, data, and use case evaluation; therefore, there is no need for domain experts or ensembles. 
    \item  While all bias estimation methods are assessing bias in one feature at a time (targeted), BENN estimates the bias for all of the features simultaneously (non-targeted).
    This enables the discovery of indirect bias in the induced ML model, i.e., discovering bias based on features that are correlated with the examined feature~\cite{mehrabi2019survey}. 
\end{itemize}

\section{\label{Background}Background}
Machine learning fairness has been addressed from various social and ethical perspectives~\cite{mehrabi2019survey}.
The most common one is group fairness~\cite{dwork2012fairness,verma2018fairness,mehrabi2019survey}, which is the absence of unethical discrimination towards any of the data distribution groups. 
For example, group fairness is present in the gender feature when men and women are treated the same by the ML model, i.e., discrimination towards one of them is not present.
When an ML model demonstrates discrimination, it might be biased towards at least one of the data subgroups, i.e., men or women. 

Several civil rights acts, such as the Fair Housing Act (FHA)\footnote{FHA:\url{hud.gov/program_offices/fair_housing_equal_opp/fair_housing_act_overview}} and the Equal Credit Opportunity Act (ECOA)\footnote{ECOA:\url{investopedia.com/terms/e/ecoa.asp}} defined several protected features, such as gender, race, skin color, national origin, religion, and marital status~\cite{mehrabi2019survey}. 
Discrimination based on the values of such protected features, as they are termed, is considered ethically unacceptable~\cite{mehrabi2019survey}.

Bias detection techniques aim to reveal underlying bias presented toward the protected feature, while bias mitigation techniques aim at reducing ML model bias~\cite{mehrabi2019survey}; there are three main types of techniques: pre-processing (adjusting training data distribution), in-processing (adjusting the ML model during training), and post-processing (adjusting the ML model's output) techniques~\cite{friedler2019comparative}.

In our experiments, we used a pre-processing technique called re-weighting mitigation~\cite{calders2009building}, which tries to achieve fairness in the training data by replicating data samples that contribute to the training set fairness.
This mitigation technique is based on optimizing the demographic parity fairness measure~\cite{dwork2012fairness}.

\section{\label{related}Related Work}
The main principle guiding bias detection methods is the ``fairness through unawareness'' principle, which can be partially represented by a statistical rule. 
Existing detection methods produce binary output by determining whether a certain statistical rule is met, and if so, the ML model will be considered fair~\cite{verma2018fairness}.

Some existing methods, such as disparate impact~\cite{feldman2015certifying} and demographic parity~\cite{dwork2012fairness}, require only the ML model predictions (i.e., minimal input).
Other methods require ground truth labels (e.g., equalized odds~\cite{hardt2016equality}, balance error rate~\cite{feldman2015certifying}, LR+ measure~\cite{feldman2015certifying}, and equal positive prediction value~\cite{berk2018fairness}); others are based on a data property called the risk score.
An example of the latter can be seen in the bank loan granting task.
The loan duration can reflect the potential risk for the bank, and therefore it can be considered a risk score.
Examples for such methods are calibration~\cite{chouldechova2017fair}, prediction parity~\cite{chouldechova2017fair}, and error rate balance with score (ERBS)~\cite{chouldechova2017fair}.

As noted, each detection method explores a different ethical aspect.
For example, sensitivity~\cite{feldman2015certifying} states that when the true positive rates (TPRs) of each protected feature value are equal, the ML model is considered fair.
While the sensitivity method aims to achieve equal TPRs, the equal accuracy~\cite{berk2018fairness} method aims at achieving equal accuracy for each protected feature value.
Both methods require the ML model predictions and ground truth as input, yet each one examines a different aspect of the ML model's fairness.
For that reason, the two methods may result in inconsistent output, i.e., the sensitivity method might determine that the examined ML model is fair and equal accuracy might not.

In addition, in order to determine which methods can be applied to the examined ML model, a domain expert is required.
For example, any detection method that requires ground truth labels, such as treatment equality~\cite{verma2018fairness} and equal false positive rate ~\cite{berk2018fairness}, cannot be applied on unsupervised ML models.

In contrast to methods aimed at the detection of bias, there are methods that produce bias estimations~\cite{vzliobaite2017measuring}, i.e., provide a number instead of a binary value. 
Examples of such methods are the normalized difference~\cite{zliobaite2015relation}, mutual information~\cite{fukuchi2015prediction}, and balance residuals~\cite{calders2013controlling} methods.
BENN's output is an estimation, and therefore it is a bias estimation method. 

Existing bias estimation methods produce estimations with different ranges and scales.
For example, the normalized difference \cite{zliobaite2015relation} method produces estimations that range between $[-1,1]$, and mutual information~\cite{fukuchi2015prediction} produces estimations that range between $[0,1]$ where zero indicates complete fairness.

As noted, the best practice for a comprehensive evaluation is to apply an ensemble of all methods, however since each method produces different output, a domain expert is required.
For example, in order to adjust the equal accuracy~\cite{berk2018fairness} method to produce a scaled bias estimation, the accuracy of each protected feature value is measured; then the accuracy variance is calculated and must be scaled using a normalization techniques such as min-max normalization.

In addition, existing methods aim at evaluating the ML model for bias based on a specific feature, which we define as \textit{targeted} evaluation.
To allow existing methods to evaluate the bias of an ML model based on all available features, a targeted evaluation needs to be performed in a brute-force manner.
This type of evaluation can be defined as a \textit{non-targeted} evaluation.
In contrast to existing methods, BENN supports both targeted and non-targeted bias evaluations in one execution.

\begin{figure*}[t]
  \centering
  \includegraphics[width=0.9\textwidth]{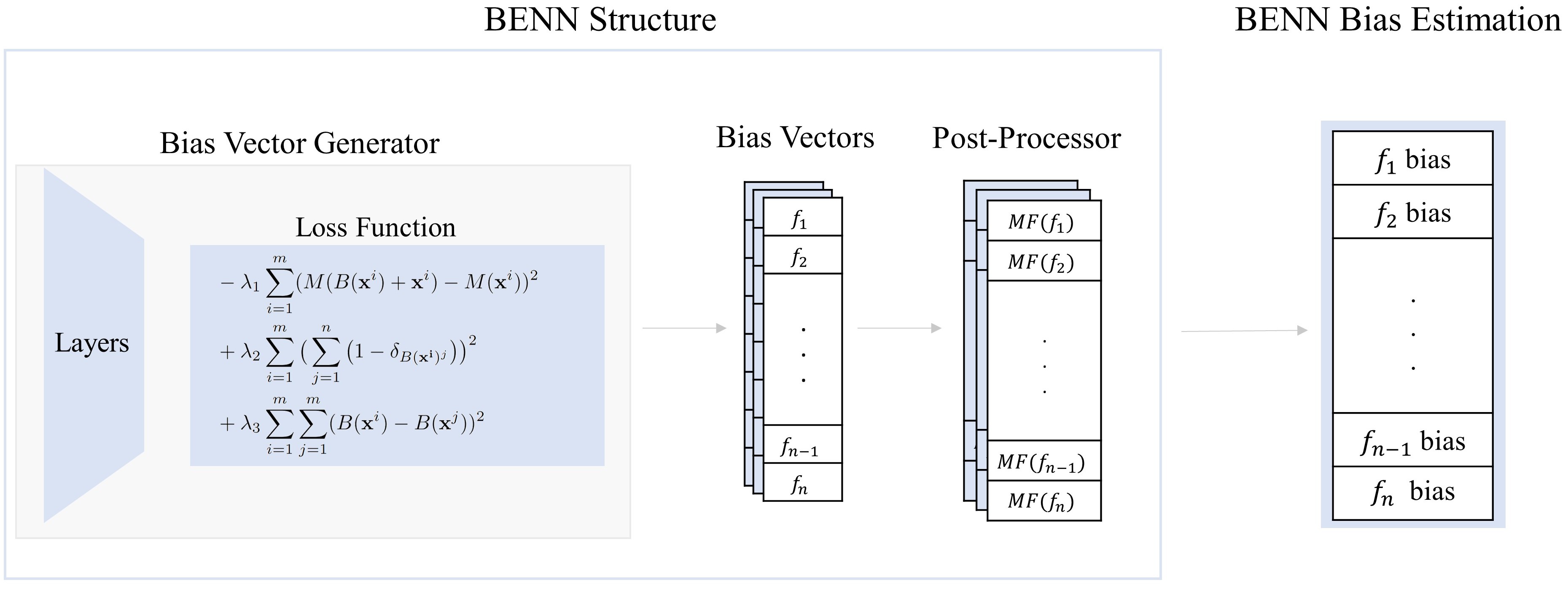}
  \caption{BENN's components and structure. The illustrated process is for non-targeted bias estimation. BENN processes the input using the bias vector generator, which produces one bias vector for each input sample. Then, the post-processor processes the bias vectors, using the mathematical aggregation $MF$, into a bias estimation for each feature.}
  \label{fig:fig1}
\end{figure*}

\section{\label{BENN_s}BENN: Bias Estimation Using DNN}
In this section, we introduce BENN's components and structure (illustrated in Figure~\ref{fig:fig1}).
First, we describe the bias vector generator, which is an unsupervised DNN with a custom loss function.
By using our custom loss function, the bias vector generator is forced to induce a hidden representation of the input data, which indicates the ML model's underlying bias for each feature.
Second, we describe the post-processor, which, given a set of bias vectors, processes them into a bias estimation for each feature.
As input, BENN receives a test set and black-box access to query the ML model examined; then BENN performs the evaluation and produces bias estimations for all of the features. 
Note that in order to perform accurate bias analysis, the test set should consist of at least one sample for all possible values for each feature examined and be sampled from the same distribution as the training set used to induce the examined ML model.

The notation used is as follows: 
Let $X\sim~D^n(FP,FU)$ be test data samples with $n$ dimensions derived from a distribution $D$, while $FP$ and $FU$ be sets of protected and unprotected features accordingly.
Let $f_p \in FP$ be a protected feature with values in $\{0,1\}$ (as is customary in the field).
Let $M$ be the ML model to be examined.
For a data sample $\mathbf{x} \in X$, let $M(\mathbf{x})$ be the $M$ outcome for $\mathbf{x}$.

\subsection{\label{BVG}Bias Vector Generator}
During the bias vector generator training, a custom loss function is used.
The custom loss function has three components which, when combined, enables the production of vectors that represent the ML model's underlying bias.

The first component of the loss function, referred to as the \textit{prediction change component}, is defined according to the fairness through unawareness~\cite{verma2018fairness} principle (i.e., the protected features should not contribute to the model's decision).
This component explores the changes that need to be made to a given sample in order to alter the given sample prediction produced by the ML model. 
This component is formalized in Equation~\ref{equation:bv_1}:
\begin{equation}
\label{equation:bv_1}
max_{B(\mathbf{x})}\left(\left|M(\mathbf{x}) - M(B(\mathbf{x})+\mathbf{x})\right|\right)
\end{equation}
where $M(\mathbf{x})$ is the model $M$'s prediction for sample $\mathbf{x}$, and $M(B(\mathbf{x})+\mathbf{x})$ is the model outcome for sample $\mathbf{x}$ and the corresponding bias vector $B(\mathbf{x})$ element-wise sum.
The \textit{prediction change component} aims at maximizing the difference between the original outcome of the ML model and the outcome after adding the bias vector.
According to the fairness through unawareness principle, in a fair ML model the protected features should have a value of zero in the corresponding bias vector entries, since they should not affect the ML model's outcome. 

However, enforcing only the \textit{prediction change component}, in an attempt to maximize the change in the ML model's outcome, may result in bias vectors with all non-zero entries.
In order to prevent this scenario, we introduce a second loss function component, referred to as the \textit{feature selection component}, which maximizes the number of entries with a zero value, i.e., minimizing the number of entries with a non-zero value.
This component is formalized in Equation~\ref{equation:bv_2}:
\begin{equation}
\label{equation:bv_2}
min_{B(\mathbf{x})}\big(\sum_{i=1}^n \big(1-\delta_{B(\mathbf{x})^{i}}\big)\big)
\end{equation}
where $B(\mathbf{x})^{i}$ is the bias vector $B(\mathbf{x})$ value in the $i$ feature, n is the number of features, and $\delta_{B(x)_i}$ is a Kronecker delta which is one if $B(x)_i = 0$ and zero if $B(x)_i\neq0$.
Accordingly, only the features that contribute most to the model decision will have non-zero values in their corresponding entries (minimal change in a minimal number of features).

However, given two different samples, the generator may produce two different vectors. 
Therefore, forcing the two previous components may cause the bias vectors produced to be significantly different.
Yet, when bias analysis is performed, the analysis should reflect all of the model decisions combined, i.e., the analysis should be performed at the feature level and not at the sample level.
The third component, referred to as the \textit{similarity component}, addresses this issue, as it enforces a minimal difference between the bias vectors, i.e., for bias vectors $B(\mathbf{x}^{i}), B(\mathbf{x}^{j})$ and a difference function $dif$, the $dif(B(\mathbf{x}^{i}), B(\mathbf{x}^{j}))$ is minimized by the loss function.
This component is formalized in Equation~\ref{equation:bv_3}:
\begin{equation}
\label{equation:bv_3}
min_{B^{i}, B^{j}}(dif(B^{i}, B^{j}))
\end{equation}
where $B_i, B_j$ are the bias vectors produced for samples $\mathbf{x}^{i}, \mathbf{x}^{j}$ correspondingly.
Accordingly, the generator is encouraged to produce similar bias vectors, which reflect the model's behavior across all model outcomes.

The complete loss function is formalized in Equation~\ref{equation:loss_complete}:
\begin{equation}
\label{equation:loss_complete}
    \begin{split}
    \mathcal{L}_{BENN} =  &
     -\lambda_1 \sum_{i=1}^m (M(B(\mathbf{x}^{i})+\mathbf{x}^{i})-M(\mathbf{x}^{i}))^2 \\
    & +\lambda_2 \sum_{i=1}^m \big(\sum_{j=1}^n \big(1-\delta_{B(\mathbf{x^{i}})^{j}}\big)\big)^2 \\
    & +\lambda_3 \sum_{i=1}^m \sum_{j=1}^m (B(\mathbf{x}^{i})-B(\mathbf{x}^{j}))^2
    \end{split}
\end{equation}
where $\mathbf{x}, \mathbf{x}_i, \mathbf{x}_j$ are samples, $\lambda_1, \lambda_2, \lambda_3$ are empirically chosen
coefficients, $\delta_{B(x)_i}$ is a Kronecker delta which is one if $B(x)_i = 0$ and zero if $B(x)_i\neq0$, $m$ is the number of vectors produced, and $B(\mathbf{x})$ is the bias vector generated according to $\mathbf{x}$.

The generator's overall goal is to minimize the loss value, which produced according to the three components described above.
The goal of the \textit{prediction change component} is to maximize the change in the model's prediction; therefore the value of this component is subtracted from the total loss value, i.e., greater model prediction change results in a smaller loss value.
The goal of the \textit{feature selection component} is to minimize the number of non-zero values in the bias vector; therefore the value of this component is added to the total loss value, i.e., a smaller number of non-zero values in the bias vector results in a smaller loss value.
The goal of the \textit{similarity component} is to minimize the difference between the bias vectors in the same training batch. 
For that reason, this component is added to the total loss value, i.e., a smaller difference between the bias vectors results in a smaller loss value.

The structure and hyperparameters of the bias vector generator are empirically chosen, as presented in Section~\ref{Experiments}.

\subsection{Post-Processor}
The main goal of the post-processor is to combine the bias vectors produced into a single vector representing the bias estimation for each feature.
The post-processor performs a mathematical aggregation by calculating the absolute average of each entry across all the bias vectors.
This aggregation is formalized in Equation~\ref{equation:post_processor}:
\begin{equation}
\label{equation:post_processor}
    \begin{split}
    post(\mathbf{b}_i)=\frac{1}{m} \sum_{j=1}^m |\mathbf{b}_i^{(j)}|
    \end{split}
\end{equation}
where $\mathbf{b}_i$ is the bias vector entry in the $i$ place, and $m$ is the number of produced vectors.
Note that in a targeted evaluation scenario, the values for the predefined protected features are extracted from the corresponding entries of the post-processor's final output.

\section{Evaluation} \label{Experiments}
\subsection{Datasets}
The following datasets were used to evaluate BENN:
\subsubsection{ProPublica COMPAS}~\cite{angwin2016machine}\footnote{github.com/propublica/compas-analysis/blob/master /compas-scores-two-years.csv} is a benchmark dataset that contains racial bias. 
The dataset was collected from the COMPAS system's historical records, which were used to assess the likelihood of a defendant to be a repeat offender. 
After filtering samples with missing values and nonmeaningful features, the dataset contains 7,215 samples and 10 features.

\subsubsection{Adult Census Income}~\cite{blake1998adult}\footnote{archive.ics.uci.edu/ml/datasets/adult} is a benchmark dataset that contains racial and gender-based bias. 
The dataset corresponds to an income level prediction task.
After filtering samples with missing values and nonmeaningful features, the dataset contains 23,562 samples and 12 features.

\subsubsection{Statlog (German Credit Data)}~\cite{kamiran2009classifying,Dua:2019}\footnote{archive.ics.uci.edu/ml/datasets/statlog+\\(german+credit+data)} is a benchmark dataset that contains gender-based bias.
The dataset corresponds to the task of determining whether the customer should qualify for a loan.
After filtering samples with missing values and nonmeaningful features, the dataset contains 1,000 samples and 20 features.

\subsubsection{Telco Churn} Additional experiments were performed on a European Telco's churn prediction ML model and dataset.
This ML model is a DNN-based model, which predicts customer churn, i.e., whether a customer will end his/her Telco subscription.
The Telco's churn dataset contains real customers information, therefore is not public.
The data contains 95,704 samples and 28 features, and the protected feature is gender.

\subsubsection{Synthetic Data} In order to provide a sanity-check, we generated a synthetic dataset. 
The dataset contains three binary features, two of which are protected: one is a fair feature (has no bias) and one is extremely biased (has maximal bias). 
The dataset consists of 305 samples, which are composed from every possible combination of the feature values.

\subsection{Ensemble Baseline}
We compared BENN's results to the results obtained by all 21 existing bias detection and estimation methods:
equalized odds~\cite{hardt2016equality}, disparate impact~\cite{feldman2015certifying}, demographic parity~\cite{dwork2012fairness}, sensitivity~\cite{feldman2015certifying}, specificity~\cite{feldman2015certifying}, balance error rate~\cite{feldman2015certifying}, LR+ measure~\cite{feldman2015certifying}, equal positive prediction value~\cite{berk2018fairness}, equal negative prediction value~\cite{berk2018fairness}, equal accuracy~\cite{berk2018fairness}, equal opportunity~\cite{hardt2016equality}, treatment equality~\cite{verma2018fairness}, equal false positive rate ~\cite{berk2018fairness}, equal false negative rate~\cite{berk2018fairness}, error rate balance~\cite{narayanan2018translation}, normalized difference~\cite{zliobaite2015relation}, mutual information~\cite{fukuchi2015prediction}, balance residuals~\cite{calders2013controlling}, calibration~\cite{chouldechova2017fair}, prediction parity~\cite{chouldechova2017fair}, and error rate balance with score (ERBS)~\cite{chouldechova2017fair}.

Due to the different outputs of the 21 existing methods, we adjusted them to produce bias estimations with the same range and scale.
The adjustments were performed according to each method's output type: binary bias detection or non-scaled bias estimation. 
In order to adjust the binary bias detection methods, we subtracted the two expressions that form the method's statistical rule and scaled the difference between the two expressions so it is between $[0,1]$ (if needed).
In the case of non-binary features (multiple values), we computed the method's statistical rule for each feature value and used the results' variance and scaled it (if needed).
In order to adjust non-scaled bias estimation methods, we scaled their outputs to be between $[0,1]$, with zero indicating complete fairness.

In order to create one estimation to which we can compare BENN's estimation, we constructed an ensemble baseline based on the 21 adjusted methods (as described above).
Each method evaluates a different ethical aspect, which may result in inconsistent estimations, therefore the ensemble baseline's estimation is set at the most severe estimation among the 21 different methods, i.e., the highest bias estimation produced for each feature.
 
Due to space limitations, we only present the ensemble baseline's final results.

\subsection{\label{Hypotheses}Evaluation Guidelines}
In order for BENN's estimations to be aligned with the ensemble's estimations, the following guidelines must be hold:

First, BENN must not overlook bias that was detected by one of the 21 existing methods.   
Therefore, for a specific feature and ML model, BENN's bias estimation should not be lower than the ensemble's estimation.
This is formalized in Expression~\ref{equation:hypothesis1}:
\begin{equation}
\label{equation:hypothesis1}
\begin{split}
{BENN}_{f_i} \geq Ensemble_{f_i}
\end{split}
\end{equation}
where $f_i$ is a specific examined feature.

Second, BENN must maintain the same ranking (order of feature estimations) as the ranking provided by the ensemble;
i.e., by ranking the features in descending order based on their estimations, BENN and the ensemble should result in an identical ranking.
This is formalized in Expression~\ref{equation:hypothesis2}:
\begin{equation}
\label{equation:hypothesis2}
\begin{split}
rank\left({BENN}_{f_i}\right) = rank\left(Ensemble_{f_i}\right)
\end{split}
\end{equation}
where $f_i$ is a specific feature, and $rank$ is the bias estimation rank. 

Third, the differences between BENN's estimations and the ensemble's estimations must be similar (close to zero variance) for all data features.
This formalized by Expression~\ref{equation:hypothesis3}:
\begin{equation}
\label{equation:hypothesis3}
\begin{split}
{BENN}_{f_i} - Ensemble_{f_i} \cong {BENN}_{f_j} - Ensemble_{f_j}
\end{split}
\end{equation}
where $f_i, f_j$ are examined features.
This ensures that the differences between BENN and the ensemble are consistent (not random) across all data features.

\begin{table*}[htb]
\centering
\begin{adjustbox}{width=1\textwidth,center}
\begin{tabular}{|l|l||c|c||c|c|c||c|c||c||c||}
 \multicolumn{2}{c||}{} & \multicolumn{2}{c||}{Synthetic data} & \multicolumn{3}{c||}{COMPAS} & \multicolumn{2}{c||}{Adult} & Statlog & Churn Prediction \\ 
 \multicolumn{2}{c||}{} & \multicolumn{1}{c}{Fair}& \multicolumn{1}{c||}{Biased} & \multicolumn{1}{c}{Race} & \multicolumn{1}{c}{Gender} & \multicolumn{1}{c||}{Age} & \multicolumn{1}{c}{Race} & \multicolumn{1}{c||}{Gender} & Gender & Gender \\ \hline 
 \multirow{2}{*}{Estimation} & Baseline & 0 & 1 & 0.4513 & 0.2955 & 0.3848 & 0.5304 & 0.6384 & 0.2215 & 0.29 \\
 & BENN &\cellcolor{Gray}0.0536 &\cellcolor{Gray}0.9948 &\cellcolor{Gray}0.662 &\cellcolor{Gray}0.5101 &\cellcolor{Gray}0.6529 &\cellcolor{Gray}0.604 &\cellcolor{Gray}0.6905 &\cellcolor{Gray}0.5293 &\cellcolor{Gray}0.5366 \\ \hline 
 \multirow{2}{*}{Rank} & Baseline & 2 & 1 & 1 & 3 & 2 & 2 & 1 & 1 & 1 \\ 
  & BENN &\cellcolor{Gray}2 &\cellcolor{Gray}1 &\cellcolor{Gray}1 &\cellcolor{Gray}3 &\cellcolor{Gray}2 &\cellcolor{Gray}2 &\cellcolor{Gray}1 &\cellcolor{Gray}1 &\cellcolor{Gray}1 \\ \hline 
 \multicolumn{2}{|l||}{Difference} & 0.0536 & -0.0052 & 0.2108 & 0.2146 & 0.2681 & 0.0737 & 0.0521 & 0.3078 & 0.2466 \\ \hline 
\multicolumn{2}{|l||}{Differences variance} & \multicolumn{2}{c||}{0.0017} & \multicolumn{3}{c||}{0.001} & \multicolumn{2}{c||}{0.0002}  & 0 & 0 \\ \hline
\end{tabular}
\end{adjustbox}
\caption{Experimental results for all use cases}\smallskip\label{resultsTable}
\end{table*}

\subsection{Experimental Settings}\label{expSettings}
All experiments were performed on the CentOS Linux 7 (Core) operating system using 24G of memory and a NVIDIA GeForce RTX 2080 Ti graphics card.
BENN and all of the code used in the experiments was written using Python 3.7.4, scikit-learn 0.22, NumPy 1.17.4, and a TensorFlow GPU 2.0.0.

The structural properties of the bias vector generator (layers parameter, optimization function, etc.) were empirically chosen and are comprised as follows:
The bias vector generator is comprised of eight dense layers with 40 units and the rectified linear unit (ReLU) as the activation function.
The output layer has the same number of units as the number of data features and hyperbolic tangent (tanh) function as the activation function.
The weight and bias initialization for every layer was randomly selected.
In order to determine the lambda parameters' values we performed experiments using each possible value in the range $[0,1]$ at steps of $0.01$ for each lambda.
As a result, the lambda values were empirically set to be equal to one.
BENN was trained using mini-batch gradient descent optimization with a batch size of 128 and 300 epochs throughout all of the experiments.
For each dataset, we induced a decision tree classifier using the scikit-learn library with the decision tree constructor default parameters. 
In order to train and evaluate the classifiers, we used five-fold cross-validation for each corresponding dataset, splitting the dataset into a training set and test set accordingly.
The test sets were used to perform the bias evaluations. 
Each experiment was performed on one pair of dataset and its corresponding ML model, which refer to as experimental use case.

As noted, in order to perform a proper bias evaluation, the test set should consist of at least one sample for each feature value examined.
For that reason, we set the seeds for different datasets differently: the seed for ProPublica COMPAS was 31, the Adult Census Income seed was 22, and the seed of the Statlog (German Credit Data) dataset was two.
Note, in the churn use case, we used an European Telco proprietary ML model, and therefore, we did not induce an additional model. 

For our experiments we defined two experimental settings: the \textit{original setting}, which uses the original dataset without any changes, and the \textit{mitigation setting}, which uses the dataset produced using the re-weighting mitigation technique~\cite{calders2009building}.
The mitigation technique parameters were set as follows: the weights of the replications with positive contribution were set at one, and the other replications weights were set at $0.1$.
In addition, the stopping criteria was defined as the probability variance threshold, which was defined as the variance of the probability for each protected feature group to obtain the positive outcome.
When the probability variance of the training set reached the probability variance threshold, the sample replication process stopped.
The variance threshold was set differently for each use case due to the difference in the initial variance: $0.0045$ for the ProPublica COMPAS and Adult Census Income datasets, $0.0003$ for the Statlog (German Credit Data) dataset, and $0.00003$ for the churn data.

\subsection{\label{Results}Experimental Results}
Table~\ref{resultsTable} presents the experimental results for the \textit{original setting} (non-mitigation setting) based on the synthetic data (fair and biased features), and the COMPAS (race, gender, and age), Adult (race and gender), Statlog (gender), and Churn prediction (gender) use cases.
For each use case, the table presents: the ensemble baseline and BENN bias estimations, the use case rankings, the difference between the estimations, and the difference variance for each protected feature.
The benchmark use case (COMPAS, Adult, Statlog) results were in the \textit{original setting} validated by five-fold cross-validation, obtaining a standard deviation below $0.02$ for every feature in every use case.

In Table~\ref{resultsTable} it can be seen that overall, BENN's estimations upheld all of the guidelines with respect to the ensemble baseline. 
With the synthetic data, both BENN and the ensemble baseline produced a bias estimation of ${\sim0}$ for the fair feature and a bias estimation of ${\sim1}$ for the biased feature.
Thus, BENN was effective in the extreme scenarios of completely fair and completely biased features correctly estimating the bias.
With the COMPAS use case, the ensemble's estimations ranged between $[\sim0.29, \sim0.45]$, and BENN's estimations ranged between $[\sim0.51, \sim0.66]$. 
All of the guidelines were upheld: BENN's estimations were higher than the ensemble's estimations for every feature; the estimation rankings were identical for the ensemble and BENN; and the difference variance was $0.001$.
With the Adult use case, the ensemble's estimations ranged between $[\sim0.53, \sim0.63]$, and BENN's estimations ranged between $[\sim0.6, \sim0.69]$. 
All of the guidelines were upheld: BENN's estimations were higher then the ensemble's estimations for every feature; the estimation rankings were identical for the ensemble baseline and BENN; and the variance was $0.0002$.
In the Statlog use case, the ensemble's estimation for the gender feature was $0.2215$, and BENN's estimation was $0.5293$; therefore, the first guideline was upheld, i.e., BENN's estimation was higher then the ensemble's estimation; there was only one protected feature, so the second and third guidelines were degenerated.
In the Churn prediction use case, the ensemble's estimation for the gender feature was $0.29$, and BENN's estimation was $0.5366$; therefore, the first guideline was upheld, i.e., BENN's estimation was higher than the ensemble's estimation; there was only one protected feature, so the second and third guidelines were degenerated.

\begin{figure*}[t]
  \centering
  \includegraphics[width=0.7\linewidth]{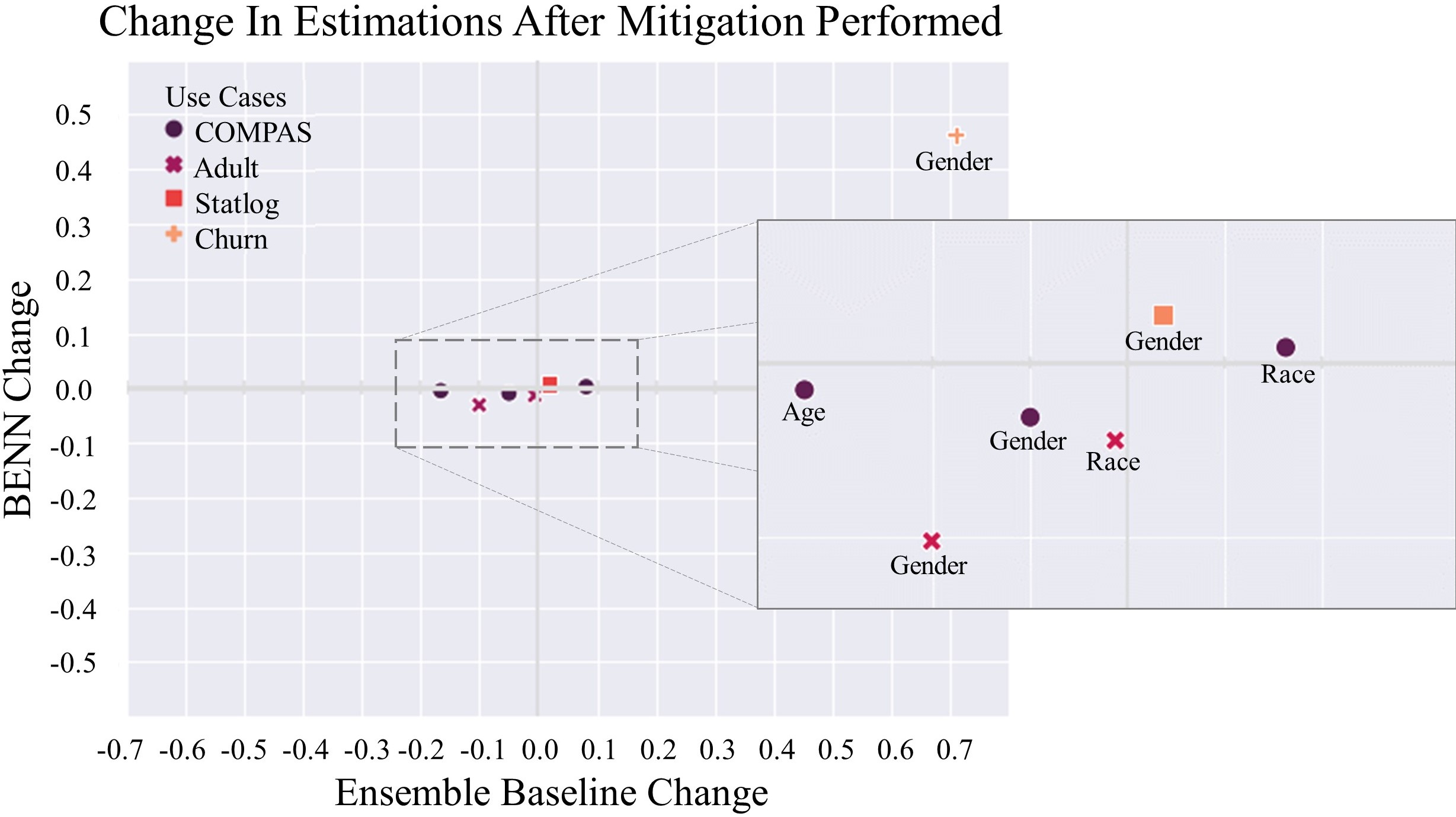}
  \caption{Mitigation experiment output. Each plotted point is an observed change of a protected feature. The x-axis is the change observed for the ensemble after the mitigation. The y-axis is the observed change for BENN after the mitigation.}
  \label{fig:fig2}
\end{figure*}

Figure~\ref{fig:fig2} presents the experimental results for the \textit{mitigation setting} based on the COMPAS (race, gender, and age features), Adult (race and gender features), Statlog (gender feature), and Churn prediction (gender feature) use cases after re-weighting mitigation was performed.
For each experiment, the charts present the change observed in BENN's estimations after the mitigation was applied (y-axis) for each corresponding change observed in the ensemble baseline (x-axis) for each use case.
The benchmark use cases results in the \textit{mitigation setting} were validated by five-fold cross-validation, with a standard deviation below $0.02$ for every feature in every use case.

Overall, we can see that BENN's estimations are similar to those of the ensemble baseline, i.e., the estimation changes are in the same direction (sign) in both cases. 
For every feature examined in every use case, a negative change in the ensemble's bias estimation corresponds with a negative change in BENN's estimation and vice versa.
Therefore, the estimations change in the same direction and exhibit similar behavior.

\section{Discussion}\label{Discussion}
In most empirical research fields in the area of artificial intelligence, when new novel methods proposed they are compared against state-of-the-art methods.
However, in the ML bias detection and estimation field, we might encounter difficulties when performing such comparisons:
\textit{i)} Outperforming existing methods is insufficient, since this is an emerging field and new methods are proposed frequently.
\textit{ii)} Each of the existing methods produces estimations differently, i.e., each method is suitable for a different use case, requires different inputs, and examines a different ethical expect.
\textit{iii)} Since each method outputs estimations of a different scale and range, one cannot simply compare their output as is typically done by using common performance measurements (accuracy, precision, etc.) 
Thus, in this paper we had to: 
\textit{i)} formulate research guidelines which BENN had to uphold to perform a field-adapted study.
\textit{ii)} create an ensemble of 21 existing methods to perform a comprehensive bias estimation.

It is worth noting that based on the experimental settings, the empirically chosen lambda parameters are exactly one for all three components: the \textit{prediction change component}, the \textit{feature selection component}, and the \textit{similarity component}.
From this one can learn that the loss function components contributes equally to the bias estimation task.
This finding emphasizes the need for all three components in bias estimation.

Furthermore, as a DNN-based solution, BENN has additional benefits, such as the ability to both learn significant patterns within the data during training and remove the dependency in the data ground truth (unsupervised learning).
However, it should be noted that using a DNN-based solution does not allow the user to understand the depth and nature of the outcomes produced (the bias estimations). 
This might be considered a disadvantage compared to existing statistical methods which have a simpler structure and more understandable estimations, i.e., have greater explainability.

Moreover, it should be noted that in this work we focused on structured data, i.e., bias estimation for unstructured data requires additional research.

\section{Conclusions and Future Work}\label{Conclusions and Future Work}
In this research, we presented BENN -- a novel method for bias estimation that uses an unsupervised DNN.
Existing methods for bias detection and estimation are limited for various reasons: \textit{i)} inconsistent and insufficient outputs made any comparison not visible, \textit{ii)} each method explores a different ethical aspect of bias \textit{iii)} each method receives different inputs.
As a result, in order to perform a comprehensive bias evaluation, a domain expert must form an ensemble of the existing methods (as we formed).
BENN is a generic method which \textit{i)} produces scaled and comprehensive bias estimations, and \textit{ii)} can be applied to any ML model without using a domain expert.
Experimental results on three benchmark datasets and one proprietary churn prediction model used by a European Telco indicate that BENN's estimations: \textit{i)} capable of revealing ML models bias, and \textit{ii)} demonstrate similar behavior to existing methods, as represented by an ensemble baseline in various settings.
Furthermore, experimental results on synthetic data indicate that BENN is capable of correctly estimating bias in extreme scenarios.
Given this, BENN can be considered a complete bias estimation technique.

Potential future work may include adapting BENN for the performance of bias evaluation in unstructured data scenarios, i.e., when the protected feature may not be explicitly presented in the data (such as image datasets).
For example, the feature \textit{gender} is not explicitly noted in face recognition image datasets, as each image is not tagged according to the gender of its subject.
In theory, utilizing object detection and classification solutions to extract the wanted feature from the data can be applied and extract the implicit feature.
In addition, making changes to the input representation can enable the extraction of a more dense representation of the input (as in the use of convolutions).
Combining both object detection and classification solutions and changing the input representation may result in an ML model and data that can be evaluated using BENN. 

Future work may also include evaluating BENN in additional experimental settings (i.e., using different datasets, additional ML algorithms, various tasks, etc.); utilizing the bias vector generator structure and outputs to perform bias mitigation; and finally, developing an explainability mechanism to support BENN structure and bias estimations.


\bibliography{references}
\newpage

\appendix
\begin{appendices}

\section{Comparison with Existing Methods}\label{BiasDetectionMethodsTable}
\begin{table*}[htb!]
\centering
\begin{adjustbox}{width=\textwidth,center}
\begin{tabular}{|p{4.5cm} p{2cm} p{2cm} p{2cm} p{2cm} p{1.7cm} p{1.4cm} p{1.8cm}|}
\hline 
Bias evaluation method & Solution level & Minimal requirements & DNN-based & Task compatibility & Scaled estimation & For non-experts & Full ethical examination\\
\hline\hline
Equalized odds \cite{hardt2016equality} & \multicolumn{1}{c}{Targeted} & \multicolumn{1}{c}{No} & \multicolumn{1}{c}{No} & \multicolumn{1}{c}{Supervised} & \multicolumn{1}{c}{No} & \multicolumn{1}{c}{No} & \multicolumn{1}{c|}{No}\\
\hline
Disparate impact \cite{feldman2015certifying} & \multicolumn{1}{c}{Targeted} & \multicolumn{1}{c}{Yes} & \multicolumn{1}{c}{No} & \multicolumn{1}{c}{Both} & \multicolumn{1}{c}{No} & \multicolumn{1}{c}{No} & \multicolumn{1}{c|}{No}\\
\hline
Demographic parity \cite{dwork2012fairness} & \multicolumn{1}{c}{Targeted} & \multicolumn{1}{c}{Yes} & \multicolumn{1}{c}{No} & \multicolumn{1}{c}{Both} & \multicolumn{1}{c}{No} & \multicolumn{1}{c}{No} & \multicolumn{1}{c|}{No}\\
\hline
Sensitivity \cite{feldman2015certifying} & \multicolumn{1}{c}{Targeted} & \multicolumn{1}{c}{No} & \multicolumn{1}{c}{No} & \multicolumn{1}{c}{Supervised} & \multicolumn{1}{c}{No} & \multicolumn{1}{c}{No} & \multicolumn{1}{c|}{No}\\
\hline
Specificity \cite{feldman2015certifying} & \multicolumn{1}{c}{Targeted} & \multicolumn{1}{c}{No} & \multicolumn{1}{c}{No} & \multicolumn{1}{c}{Supervised} & \multicolumn{1}{c}{No} & \multicolumn{1}{c}{No} & \multicolumn{1}{c|}{No}\\
\hline
Balance error rate \cite{feldman2015certifying} & \multicolumn{1}{c}{Targeted} & \multicolumn{1}{c}{No} & \multicolumn{1}{c}{No} & \multicolumn{1}{c}{Supervised} & \multicolumn{1}{c}{No} & \multicolumn{1}{c}{No} & \multicolumn{1}{c|}{No}\\
\hline
LR+ measure \cite{feldman2015certifying} & \multicolumn{1}{c}{Targeted} & \multicolumn{1}{c}{No} & \multicolumn{1}{c}{No} & \multicolumn{1}{c}{Supervised} & \multicolumn{1}{c}{No} & \multicolumn{1}{c}{No} & \multicolumn{1}{c|}{No}\\
\hline
Equal positive prediction value \cite{berk2018fairness} & \multicolumn{1}{c}{Targeted} & \multicolumn{1}{c}{No} & \multicolumn{1}{c}{No} & \multicolumn{1}{c}{Supervised} & \multicolumn{1}{c}{No} & \multicolumn{1}{c}{No} & \multicolumn{1}{c|}{No}\\
\hline
Equal negative prediction value \cite{berk2018fairness} & \multicolumn{1}{c}{Targeted} & \multicolumn{1}{c}{No} & \multicolumn{1}{c}{No} & \multicolumn{1}{c}{Supervised} & \multicolumn{1}{c}{No} & \multicolumn{1}{c}{No} & \multicolumn{1}{c|}{No}\\
\hline
Equal accuracy \cite{berk2018fairness} & \multicolumn{1}{c}{Targeted} & \multicolumn{1}{c}{No} & \multicolumn{1}{c}{No} & \multicolumn{1}{c}{Supervised} & \multicolumn{1}{c}{No} & \multicolumn{1}{c}{No} & \multicolumn{1}{c|}{No}\\
\hline
Equal opportunity \cite{hardt2016equality} & \multicolumn{1}{c}{Targeted} & \multicolumn{1}{c}{No} & \multicolumn{1}{c}{No} & \multicolumn{1}{c}{Supervised} & \multicolumn{1}{c}{No} & \multicolumn{1}{c}{No} & \multicolumn{1}{c|}{No}\\
\hline
Treatment equality \cite{verma2018fairness} & \multicolumn{1}{c}{Targeted} & \multicolumn{1}{c}{No} & \multicolumn{1}{c}{No} & \multicolumn{1}{c}{Supervised} & \multicolumn{1}{c}{No} & \multicolumn{1}{c}{No} & \multicolumn{1}{c|}{No}\\
\hline
Equal false positive rate  \cite{berk2018fairness} & \multicolumn{1}{c}{Targeted} & \multicolumn{1}{c}{No} & \multicolumn{1}{c}{No} & \multicolumn{1}{c}{Supervised} & \multicolumn{1}{c}{No} & \multicolumn{1}{c}{No} & \multicolumn{1}{c|}{No}\\
\hline
Equal false negative rate \cite{berk2018fairness} & \multicolumn{1}{c}{Targeted} & \multicolumn{1}{c}{No} & \multicolumn{1}{c}{No} & \multicolumn{1}{c}{Supervised} & \multicolumn{1}{c}{No} & \multicolumn{1}{c}{No} & \multicolumn{1}{c|}{No}\\
\hline
Error rate balance \cite{narayanan2018translation} & \multicolumn{1}{c}{Targeted} & \multicolumn{1}{c}{No} & \multicolumn{1}{c}{No} & \multicolumn{1}{c}{Supervised} & \multicolumn{1}{c}{No} & \multicolumn{1}{c}{No} & \multicolumn{1}{c|}{No}\\
\hline
Normalized difference \cite{zliobaite2015relation} & \multicolumn{1}{c}{Targeted} & \multicolumn{1}{c}{Yes} & \multicolumn{1}{c}{No} & \multicolumn{1}{c}{Both} & \multicolumn{1}{c}{No} & \multicolumn{1}{c}{No} & \multicolumn{1}{c|}{No}\\
\hline
Mutual information \cite{fukuchi2015prediction} & \multicolumn{1}{c}{Targeted} & \multicolumn{1}{c}{Yes} & \multicolumn{1}{c}{No} & \multicolumn{1}{c}{Both} & \multicolumn{1}{c}{No} & \multicolumn{1}{c}{No} & \multicolumn{1}{c|}{No}\\
\hline
Balance residuals \cite{calders2013controlling} & \multicolumn{1}{c}{Targeted} & \multicolumn{1}{c}{No} & \multicolumn{1}{c}{No} & \multicolumn{1}{c}{Supervised} & \multicolumn{1}{c}{No} & \multicolumn{1}{c}{No} & \multicolumn{1}{c|}{No}\\
\hline
Calibration \cite{chouldechova2017fair} & \multicolumn{1}{c}{Targeted} & \multicolumn{1}{c}{No} & \multicolumn{1}{c}{No} & \multicolumn{1}{c}{Both} & \multicolumn{1}{c}{No} & \multicolumn{1}{c}{No} & \multicolumn{1}{c|}{No}\\
\hline
Prediction parity \cite{chouldechova2017fair} & \multicolumn{1}{c}{Targeted} & \multicolumn{1}{c}{No} & \multicolumn{1}{c}{No} & \multicolumn{1}{c}{Both} & \multicolumn{1}{c}{No} & \multicolumn{1}{c}{No} & \multicolumn{1}{c|}{No}\\
\hline
Error rate balance with score (ERBS) \cite{chouldechova2017fair} & \multicolumn{1}{c}{Targeted} & \multicolumn{1}{c}{No} & \multicolumn{1}{c}{No} & \multicolumn{1}{c}{Both} & \multicolumn{1}{c}{No} & \multicolumn{1}{c}{No} & \multicolumn{1}{c|}{No}\\
\hline
\textbf{BENN} & \multicolumn{1}{c}{\textbf{Both}} & \multicolumn{1}{c}{\textbf{Yes}} & \multicolumn{1}{c}{\textbf{Yes}} & \multicolumn{1}{c}{\textbf{Both}} & \multicolumn{1}{c}{\textbf{Yes}} & \multicolumn{1}{c}{\textbf{Yes}} & \multicolumn{1}{c|}{\textbf{Yes}}\\
\hline
\end{tabular}
\end{adjustbox}
\caption{Existing bias detection and estimation methods and an overview of their properties}\smallskip\label{methodsTable}
\end{table*}

Table~\ref{methodsTable} provides an overview of the existing bias detection and estimation methods with respect to various properties. 

The first property is the ability to provide two different evaluation solutions: \textit{i)} evaluate the machine learning (ML) model for bias based on a specific feature, which we define as \textit{targeted} evaluation, and  \textit{ii)} evaluate the ML model for bias based on all available features, which we define as \textit{non-targeted} evaluation.
As seen in Table~\ref{methodsTable}, none of the existing methods support non-targeted bias evaluation, i.e., in order to obtain a \textit{non-targeted} evaluation, a \textit{targeted} evaluation needs to be performed in a brute-force manner.
In contrast, BENN offers both targeted and non-targeted bias evaluation solutions in one exaction.

The second property is the ability to evaluate the ML model with limited access to information, i.e., just the ML model predictions. 
Table~\ref{methodsTable} shows that the majority of the methods require additional information, such as the ground truth labels, the data's predefined risk score, etc.
In contrast, BENN requires only the model predictions.

The third property is the method's technical novelty.
As seen in the table, all of the existing methods are based on statistical rules whereas BENN is based on an unsupervised deep neural network (DNN).
As such, BENN provides a novel DNN-based bias evaluation that produces more well-founded bias estimations.

The fourth property is the method's ability to perform bias evaluation for both supervised and unsupervised ML models. 
In Table~\ref{methodsTable} it can be seen that most of the methods are only suited for supervised learning due to their dependence on ground truth labels. 
In contrast, BENN supports both supervised and unsupervised learning as it considers only the model predictions.

The fifth property is the method's ability to produce scaled bias estimations.
Most of the methods require adaptation in order to produce bias estimations, as they are aimed solely at performing bias detection. 
While there are methods that can estimate bias, they produce bias estimations with different ranges and scales, making it difficult to compare their outputs. 
In contrast, BENN is capable of producing a scaled bias estimation for each of the examined features/models.

The sixth property is the method's accessibility. 
Prior knowledge in the following fields is required in order to evaluate bias using the existing methods: ML, bias detection, bias estimation, data complexity, and domain constraints. 
Therefore, existing methods are not suitable for non-experts. 
To the best of our knowledge, the first and only attempt at making the bias evaluation process more accessible to the non-expert community was made by Myers \etal who introduced a visualization tool which aimed at discovering bias, referred to as CEB~\cite{myers2020revealing}.
However, CEB is just meant for visualization purposes and is only suitable for neural network (NN) based ML models~\cite{myers2020revealing}.
CEB visualizes the NN activation's behavior and requires additional processing in order to perform manual bias evaluation (behavior analysis, etc).
In contrast, BENN is both more accessible and better suited to the non-expert community.

Another property is the method's ability to provide a complete ethical evaluation. 
As noted, all bias detection and estimation methods are based on the ``fairness through unawareness" principle~\cite{verma2018fairness}, which means that changes in features with ethical significance should not change the ML model's outcome.
Each existing method examines only one ethical aspect of this principle, and none of them examine the entire principle, i.e., none of them perform a complete ethical evaluation.
In contrast, we empirically show that BENN performs a complete bias evaluation, as it examines the ML model for all ethical aspects derived from the ``fairness through unawareness" principle~\cite{verma2018fairness}.

Table~\ref{methodsTable} highlights the disadvantages and limitations of the existing methods for bias evaluation, pointing to the need for a generic, complete, and accessible bias evaluation technique like BENN.

\section{The Ensemble Baseline}\label{app_baseline}
In this section, we describe how the ensemble baseline was created. 
As noted, the ensemble baseline is constructed using the 21 existing bias detection and estimation methods.
Due to the different outputs of these methods, we adjusted them to produce bias estimations with the same range and scale.
The adjustments were performed according to each method's output type: binary bias detection or non-scaled bias estimation.

Existing detection methods produce binary output by determining whether a certain statistical rule is met, and if so, the ML model is considered fair~\cite{verma2018fairness}
In order to adjust the binary bias detection methods, we subtracted the two expressions that form the method's statistical rule and scaled the difference between the two expressions so it is between $[0,1]$ (if needed).
In the case of non-binary features (multiple values), we computed the method's statistical rule for each feature value and used the results' variance and scaled it (if needed).
In order to adjust non-scaled bias estimation methods, we scaled their outputs to be between $[0,1]$, with zero indicating complete fairness.

In order to create one estimation to which we can compare BENN's estimation, we constructed an ensemble baseline based on the 21 adjusted methods (as described above).
Each method evaluates a different ethical aspect, therefore the ensemble baseline's estimation is set at the most restrictive estimation prodused by the 21 different methods, i.e., the highest bias estimation produced for each feature.

In this section, we discuss the 21 existing methods and explain how we adjusted them for a scaled bias estimation.
The notation used is as follows: 
Let $X\sim D^n(FP,FU)$ be test data samples with $n$ dimensions, derived from distribution $D$, and $FP$ and $FU$ be sets of protected and unprotected features accordingly.
Let $f_p \in FP$ be a protected feature with values in $\{0,1\}$ (as is customary in the field).
Let $M$ be the ML model to be examined.
For a data sample $\mathbf{x} \in X$, let $M(\mathbf{x})$ be the model $M$ outcome for $\mathbf{x}$ and $y_t$ be $\mathbf{x}$ ground truth.
Let $C$ be a group of possible classes and $c_i\in C$ be a possible class.
Let $S(\mathbf{x})$ be sample $\mathbf{x}$'s risk score. 

\subsection{Equalized Odds}
The equalized odds method, presented in Equation~\ref{equation:equalized_odds1}, produces a binary bias detection output, therefore we processed it accordingly (as described at the beginning of this section).
\begin{equation}
\label{equation:equalized_odds1}
\begin{split}
& P(M(\mathbf{x})=1|f_p=0,y_t=c_i)\\
& = P(M(\mathbf{x})=1|f_p=1,y_t=c_i)
\end{split}
\end{equation}
In order to produce a scaled bias estimation, we performed the calculations in Equation~\ref{equation:equalized_odds2} for the protected feature values:
\begin{equation}
\label{equation:equalized_odds2}
\begin{split}
& \forall c_i \in C  ,\forall v_p \in f_p \quad Estimation=\\
& MAX\left(variance \left(p \left(M \left(\mathbf{x} \right)=1|f_p=v_p,y_t=c_i \right)\right)\right)
\end{split}
\end{equation}

\subsection{Disparate Impact}
The disparate impact method, presented in Equation~\ref{equation:disparate_impact1}, produces a non-scaled bias estimation output, therefore we processed it accordingly (as described at the beginning of this section). 
\begin{equation}
\label{equation:disparate_impact1}
\begin{split}
 DI =  \frac{P(M(\mathbf{x})=1|f_p=1)}{P(M(\mathbf{x})=1|f_p=0)}
\end{split}
\end{equation}
In order to produce a scaled bias estimation, we performed the calculations in Equation~\ref{equation:disparate_impact2} for the protected feature values:
\begin{equation}
\label{equation:disparate_impact2}
\begin{split}
 Estimation =\begin{cases}0 & min(DI)>0.8\\1-\frac{min(DI)}{0.8} & min(DI)  \leq0.8\end{cases} 
\end{split}
\end{equation}

\subsection{Demographic parity}
The demographic parity method, presented in Equation~\ref{equation:demographic_parity1}, produces a binary bias detection output, therefore we processed it accordingly (as described at the beginning of this section). 
\begin{equation}
\label{equation:demographic_parity1}
\begin{split}
 P(M(\mathbf{x})=1|f_p=1) = P(M(\mathbf{x})=1|f_p=0)
\end{split}
\end{equation}
In order to produce a scaled bias estimation, we performed the calculations in Equation~\ref{equation:demographic_parity2} for the protected feature values:
\begin{equation}
\label{equation:demographic_parity2}
\begin{split}
 Estimation = & max(|P(M(\mathbf{x})=1|f_p=1) - \\
 & P(M(\mathbf{x})=1|f_p=0)|) 
\end{split}
\end{equation}

\subsection{Sensitivity}
The sensitivity method, presented in Equation~\ref{equation:Sensitivity1}, produces a binary bias detection output, therefore we processed it accordingly (as described at the beginning of this section).
\begin{equation}
\label{equation:Sensitivity1}
\begin{split}
\frac{TP_{f_p=1}}{TP_{f_p=1} + FN_{f_p=1}} = 
\frac{TP_{f_p=0}}{TP_{f_p=0} + FN_{f_p=0}}
\end{split}
\end{equation}
In order to produce a scaled bias estimation, we performed the calculations in Equation~\ref{equation:Sensitivity2} for the protected feature values:
\begin{equation}
\label{equation:Sensitivity2}
\begin{split}
 & Estimation = \\ & max\left(\left|\frac{TP_{f_p=1}}{TP_{f_p=1} + FN_{f_p=1}} - 
\frac{TP_{f_p=0}}{TP_{f_p=0} + FN_{f_p=0}}\right|\right)
\end{split}
\end{equation}

\subsection{Specificity}
The specificity method, presented in Equation~\ref{equation:Specificity1}, produces a binary bias detection output, therefore we processed it accordingly (as described at the beginning of this section).
\begin{equation}
\label{equation:Specificity1}
\begin{split}
\frac{TN_{f_p=1}}{TN_{f_p=1} + FP_{f_p=1}} = 
\frac{TN_{f_p=0}}{TN_{f_p=0} + FP_{f_p=0}}
\end{split}
\end{equation}
In order to produce a scaled bias estimation, we performed the calculations in Equation~\ref{equation:Specificity2} for the protected feature values:
\begin{equation}
\label{equation:Specificity2}
\begin{split}
 & Estimation = \\ & max\left(\left|\frac{TN_{f_p=1}}{TN_{f_p=1} + FP_{f_p=1}} - 
\frac{TN_{f_p=0}}{TN_{f_p=0} + FP_{f_p=0}}\right|\right)
\end{split}
\end{equation}

\subsection{Balance Error Rate}
The balance error rate method, presented in Equation~\ref{equation:BER1}, produces a binary bias detection output, therefore we processed it accordingly (as described at the beginning of this section).
\begin{equation}
\label{equation:BER1}
\begin{split}
\frac{FP_{f_p=1} + FN_{f_p=1}}{2} = 
\frac{FP_{f_p=0} + FN_{f_p=0}}{2}
\end{split}
\end{equation}
In order to produce a scaled bias estimation, we performed the calculations in Equation~\ref{equation:BER2} for the protected feature values:
\begin{equation}
\label{equation:BER2}
\begin{split}
 & Estimation = \\ & \frac{max\left(\left|\frac{FP_{f_p=1} + FN_{f_p=1}}{2} - 
\frac{FP_{f_p=0} + FN_{f_p=0}}{2}\right|\right)}{\frac{data\_size}{2}}
\end{split}
\end{equation}

\subsection{LR+ Measure}
The LR+ measure method, presented in Equation~\ref{equation:LR1}, produces a binary bias detection output, therefore we processed it accordingly (as described at the beginning of this section).
\begin{equation}
\label{equation:LR1}
\begin{split}
\frac{\frac{TP_{f_p=1}}{TP_{f_p=1} + FN_{f_p=1}}}{1-\frac{TP_{f_p=1}}{TP_{f_p=1} + FN_{f_p=1}}} = 
\frac{\frac{TP_{f_p=0}}{TP_{f_p=0} + FN_{f_p=0}}}{1-\frac{TP_{f_p=0}}{TP_{f_p=0} + FN_{f_p=0}}}
\end{split}
\end{equation}
In order to produce a scaled bias estimation, we performed the calculations in Equation~\ref{equation:LR2} for the protected feature values:
\begin{equation}
\label{equation:LR2}
\begin{split}
 & Estimation = \\ & \frac{max\left(\left|\frac{\frac{TP_{f_p=1}}{TP_{f_p=1} + FN_{f_p=1}}}{1-\frac{TP_{f_p=1}}{TP_{f_p=1} + FN_{f_p=1}}} - 
\frac{\frac{TP_{f_p=0}}{TP_{f_p=0} + FN_{f_p=0}}}{1-\frac{TP_{f_p=0}}{TP_{f_p=0} + FN_{f_p=0}}}\right|\right)}{\frac{data\_size}{2}}
\end{split}
\end{equation}

\subsection{Equal Positive Prediction Value}
The equal positive prediction value method, presented in Equation~\ref{equation:eppv1}, produces a binary bias detection output, therefore we processed it accordingly (as described at the beginning of this section).
\begin{equation}
\label{equation:eppv1}
\begin{split}
\frac{TP_{f_p=1}}{TP_{f_p=1} + FP_{f_p=1}} = 
\frac{TP_{f_p=0}}{TP_{f_p=0} + FP_{f_p=0}}
\end{split}
\end{equation}
In order to produce a scaled bias estimation, we performed the calculations in Equation~\ref{equation:eppv2} for the protected feature values:
\begin{equation}
\label{equation:eppv2}
\begin{split}
 & Estimation = \\ & max\left(\left|\frac{TP_{f_p=1}}{TP_{f_p=1} + FP_{f_p=1}} -
\frac{TP_{f_p=0}}{TP_{f_p=0} + FP_{f_p=0}}\right|\right) 
\end{split}
\end{equation}

\subsection{Equal Negative Prediction Value}
The equal negative prediction Value method, presented in Equation~\ref{equation:enpv1},  produces a binary bias detection output, therefore we processed it accordingly (as described at the beginning of this section).
\begin{equation}
\label{equation:enpv1}
\begin{split}
\frac{TN_{f_p=1}}{TN_{f_p=1} + FN_{f_p=1}} = 
\frac{TN_{f_p=0}}{TN_{f_p=0} + FN_{f_p=0}}
\end{split}
\end{equation}
In order to produce a scaled bias estimation, we performed the calculations in Equation~\ref{equation:enpv2} for the protected feature values:
\begin{equation}
\label{equation:enpv2}
\begin{split}
 & Estimation = \\ & max\left(\left|\frac{TN_{f_p=1}}{TN_{f_p=1} + FN_{f_p=1}} -
\frac{TN_{f_p=0}}{TN_{f_p=0} + FN_{f_p=0}}\right|\right) 
\end{split}
\end{equation} 

\subsection{Equal Accuracy}
The equal accuracy method, presented in Equation~\ref{equation:acc1}, produces a binary bias detection output, therefore we processed it accordingly (as described at the beginning of this section).
\begin{equation}
\label{equation:acc1}
\begin{split}
\frac{TN_{f_p=1} + TP_{f_p=1}}{data\_size} = 
\frac{TN_{f_p=0} + TP_{f_p=0}}{data\_size}
\end{split}
\end{equation}
In order to produce a scaled bias estimation, we performed the calculations in Equation~\ref{equation:acc2} for the protected feature values:
\begin{equation}
\label{equation:acc2}
\begin{split}
 & Estimation = \\ & max\left(\left|\frac{TN_{f_p=1} + TP_{f_p=1}}{data\_size} -
\frac{TN_{f_p=0} + TP_{f_p=0}}{data\_size}\right|\right) 
\end{split}
\end{equation}

\subsection{Equal Opportunity}
The equal opportunity method, presented in Equation~\ref{equation:eqopp1}, produces a binary bias detection output, therefore we processed it accordingly (as described at the beginning of this section).
\begin{equation}
\label{equation:eqopp1}
\begin{split}
& P(M(\mathbf{x})=1|f_p=0,y_t=1)\\
& = P(M(\mathbf{x})=1|f_p=1,y_t=1)
\end{split}
\end{equation}
In order to produce a scaled bias estimation, we performed the calculations in Equation~\ref{equation:eqopp2} for the protected feature values:
\begin{equation}
\label{equation:eqopp2}
\begin{split}
 & Estimation = \\ & 1- variance \left(p \left(M \left(\mathbf{x} \right)=1|f_p=v_p,y_t=c_i \right)\right)
\end{split}
\end{equation}

\subsection{Treatment Equality}
The treatment equality method, presented in Equation~\ref{equation:te1}, produces a binary bias detection output, therefore we processed it accordingly (as described at the beginning of this section).
\begin{equation}
\label{equation:te1}
\begin{split}
\frac{FN_{f_p=1}}{FP_{f_p=1}} = 
\frac{FN_{f_p=0}}{FP_{f_p=0}}
\end{split}
\end{equation}
In order to produce a scaled bias estimation, we performed the calculations in Equation~\ref{equation:te2} for the protected feature values:
\begin{equation}
\label{equation:te2}
\begin{split}
 Estimation = \frac{max\left(\left|\frac{FN_{f_p=1}}{FP_{f_p=1}} -
\frac{FN_{f_p=0}}{FP_{f_p=0}}\right|\right)}{data\_size}  
\end{split}
\end{equation}

\subsection{Equal False Positive Rate}
The equal false positive rate method, presented in Equation~\ref{equation:efpr1}, produces a binary bias detection output, therefore we processed it accordingly (as described at the beginning of this section).
\begin{equation}
\label{equation:efpr1}
\begin{split}
\frac{FP_{f_p=1}}{FP_{f_p=1} + TN_{f_p=1}} = 
\frac{FP_{f_p=0}}{FP_{f_p=0} + TN_{f_p=0}}
\end{split}
\end{equation}
In order to produce a scaled bias estimation, we performed the calculations in Equation~\ref{equation:efpr2} for the protected feature values:
\begin{equation}
\label{equation:efpr2}
\begin{split}
 & Estimation = \\ & max\left(\left|\frac{FP_{f_p=1}}{FP_{f_p=1} + TN_{f_p=1}} -
\frac{FP_{f_p=0}}{FP_{f_p=0} + TN_{f_p=0}}\right|\right) 
\end{split}
\end{equation}

\subsection{Equal False Negative Rate}
The equal false negative rate method, presented in Equation~\ref{equation:efnr1}, produces a binary bias detection output, therefore we processed it accordingly (as described at the beginning of this section).
\begin{equation}
\label{equation:efnr1}
\begin{split}
\frac{FN_{f_p=1}}{FN_{f_p=1} + TP_{f_p=1}} = 
\frac{FN_{f_p=0}}{FN_{f_p=0} + TP_{f_p=0}}
\end{split}
\end{equation}
In order to produce a scaled bias estimation, we performed the calculations in Equation~\ref{equation:efnr2} for the protected feature values:
\begin{equation}
\label{equation:efnr2}
\begin{split}
 & Estimation = \\ & max\left(\left|\frac{FN_{f_p=1}}{FN_{f_p=1} + TP_{f_p=1}} -
\frac{FN_{f_p=0}}{FN_{f_p=0} + TP_{f_p=0}}\right|\right) 
\end{split}
\end{equation}

\subsection{Error Rate Balance}
The error rate balance method, presented in Equation~\ref{equation:erb1}, produces a binary bias detection output, therefore we processed it accordingly (as described at the beginning of this section).
\begin{equation}
\label{equation:erb1}
\begin{split}
& \frac{FN_{f_p=1}}{FN_{f_p=1} + TP_{f_p=1}} = 
\frac{FN_{f_p=0}}{FN_{f_p=0} + TP_{f_p=0}}\\
& \qquad\qquad\qquad\qquad and\\
& \frac{FP_{f_p=1}}{FP_{f_p=1} + TN_{f_p=1}} = 
\frac{FP_{f_p=0}}{FP_{f_p=0} + TN_{f_p=0}}
\end{split}
\end{equation}
In order to produce a scaled bias estimation, we performed the calculations in Equation~\ref{equation:erb2} for the protected feature values:
\begin{equation}
\label{equation:erb2}
\begin{split}
 & Estimation = \\ & 1-min\left(\text{\textit{Equal FNR Estimation}},\text{\textit{Equal FPR Estimation}} \right) 
\end{split}
\end{equation}

\subsection{Normalized Difference}
The normalized difference method, presented in Equation~\ref{equation:nd1}, produces a non-scaled bias estimation output, therefore we processed it accordingly (as described at the beginning of this section).
\begin{equation}
\label{equation:nd1}
\begin{split}
 ND =  \frac{P(M(\mathbf{x})=1|f_p=1)-P(M(\mathbf{x})=1|f_p=1)}
 {max\left(\frac{P(M(\mathbf{x})=1)}{p(f_p=0)},\frac{P(M(\mathbf{x})=0)}{p(f_p=1)} \right)}
\end{split}
\end{equation}
In order to produce a scaled bias estimation, we performed the calculations in Equation~\ref{equation:nd2} for the protected feature values:
\begin{equation}
\label{equation:nd2}
\begin{split}
 Estimation = max\left(\left|ND \right| \right) 
\end{split}
\end{equation}

\subsection{Mutual Information}
The mutual information method, presented in Equation~\ref{equation:mi1}, produces a non-scaled bias estimation output, therefore we processed it accordingly (as described at the beginning of this section).
\begin{equation}
\label{equation:mi1}
\begin{split}
 & MI =  \frac{I(M(\mathbf{x}),f_p)}{\sqrt{H(M(\mathbf{x}))H(f_p)}}\\
 & I(M(\mathbf{x}),f_p) =\\
 & \sum_{M(\mathbf{x}),f_p} P(M(\mathbf{x}),f_p)log\left(\frac{P(M(\mathbf{x}),f_p)}{P(M(\mathbf{x}))P(f_p)} \right)\\
 & H(a) = -\sum_a P(a)log(p(a))
\end{split}
\end{equation}
In order to produce a scaled bias estimation, we performed the calculations in Equation~\ref{equation:mi2} for the protected feature values:
\begin{equation}
\label{equation:mi2}
\begin{split}
 Estimation = max\left(\left|MI \right| \right) 
\end{split}
\end{equation}

\subsection{Balance Residuals} 
The balance residuals method, presented in Equation~\ref{equation:br1}, produces a non-scaled bias estimation output, therefore we processed it accordingly (as described at the beginning of this section).
\begin{equation}
\label{equation:br1}
\begin{split}
 & BR =  \frac{\sum_{f_p=1} \left| y_t - M(\mathbf{x})\right|}{\left| f_p=1\right|} - 
 \frac{\sum_{f_p=0} \left| y_t - M(\mathbf{x})\right|}{\left| f_p=0\right|}
\end{split}
\end{equation}
In order to produce a scaled bias estimation, we performed the calculations in Equation~\ref{equation:br2} for the protected feature values:
\begin{equation}
\label{equation:br2}
\begin{split}
 Estimation = max\left(\left|BR \right| \right) 
\end{split}
\end{equation}

\subsection{Calibration}
The calibration method, presented in Equation~\ref{equation:Calibration1}, produces a binary bias detection output, therefore we processed it accordingly (as described at the beginning of this section).
\begin{equation}
\label{equation:Calibration1}
\begin{split}
P(M(\mathbf{x})=1|s(\mathbf{x}), f_p=1) = P(M(\mathbf{x})=1|s(\mathbf{x}), f_p=0)
\end{split}
\end{equation}
In order to produce a scaled bias estimation, we performed the calculations in Equation~\ref{equation:Calibration2} for the protected feature values:
\begin{equation}
\label{equation:Calibration2}
\begin{split}
 & Estimation = \\ & min\left(variance \left(P(M(\mathbf{x})=1|s(\mathbf{x}), f_p=v_p)\right)\right)
\end{split}
\end{equation}

\subsection{Prediction Parity}
The prediction parity method, presented in Equation~\ref{equation:pp1}, produces a binary bias detection output, therefore we processed it accordingly (as described at the beginning of this section).
\begin{equation}
\label{equation:pp1}
\begin{split}
& P(M(\mathbf{x})=1|s(\mathbf{x})>t, f_p=1) =\\
& P(M(\mathbf{x})=1|s(\mathbf{x})>t, f_p=0)
\end{split}
\end{equation}
where $t$ is a given risk score threshold.
In order to produce a scaled bias estimation, we performed the calculations in Equation~\ref{equation:pp2} for the protected feature values:
\begin{equation}
\label{equation:pp2}
\begin{split}
 & Estimation = \\ & 1-variance \left(P(M(\mathbf{x})=1|s(\mathbf{x})>t, f_p=v_p)\right)
\end{split}
\end{equation}

\subsection{Error Rate Balance with Score}
The error rate balance with score method, presented in Equation~\ref{equation:erbs1}, produces a binary bias detection output, therefore we processed it accordingly (as described at the beginning of this section).
\begin{equation}
\label{equation:erbs1}
\begin{split}
& P(s(\mathbf{x})>t| M(\mathbf{x})=0, f_p=1) =\\
& P(s(\mathbf{x})>t| M(\mathbf{x})=0, f_p=0)\\
& \qquad\qquad\qquad and\\
& P(s(\mathbf{x})\leq t| M(\mathbf{x})=1, f_p=1) =\\
& P(s(\mathbf{x})\leq t| M(\mathbf{x})=1, f_p=0)
\end{split}
\end{equation}
where $t$ is a given risk score threshold.
In order to produce a scaled bias estimation, we performed the calculations in Equation~\ref{equation:erbs2} for the protected feature values:
\begin{equation}
\label{equation:erbs2}
\begin{split}
 & Estimation = \\ 
 & 1- min(variance(P(s(\mathbf{x})>t| M(\mathbf{x})=0, f_p=v_p)), \\
 & variance(P(s(\mathbf{x})\leq t| M(\mathbf{x})=1, f_p=v_p)))
\end{split}
\end{equation}

\section{Ethical Aspects of the Compared Methods }
\begin{table*}[htb!]
\centering
\begin{adjustbox}{width=\textwidth,center}
\begin{tabular}{|p{4.5cm} p{15cm} |}
\hline 
Bias evaluation method & Ethical aspect examined\\
\hline\hline
Equalized odds \cite{hardt2016equality} & Given a specific ground truth label, examines whether the protected feature value affects the ML model's probability of producing a positive ML model's outcome.\\
\hline
Disparate impact \cite{feldman2015certifying} & Examines whether the protected feature value affects the ML model's probability of producing a positive ML model's outcome.\\
\hline
Demographic parity \cite{dwork2012fairness} & Examines whether the protected feature value affects the ML model's probability of producing a positive ML model's outcome.\\
\hline
Sensitivity \cite{feldman2015certifying} & Examines whether the protected feature value affects the true positive rate (TPR) of the ML model's outcomes.\\
\hline
Specificity \cite{feldman2015certifying} & Examines whether the protected feature value affects the true negative rate (TNR) of the ML model's outcomes.\\
\hline
Balance error rate \cite{feldman2015certifying} & Examines whether the protected feature value affects the error level of the ML model's outcomes.\\
\hline
LR+ measure \cite{feldman2015certifying} & Checks whether the protected feature value affects the TPR opposite ratio of the ML model's outcomes.\\
\hline
Equal positive prediction value \cite{berk2018fairness} & Examines whether the protected feature value affects the precision of the ML model's outcomes.\\
\hline
Equal negative prediction value \cite{berk2018fairness} & Examines whether the protected feature value affects the ratio between the number of true positive (TP) and negative outcomes.\\
\hline
Equal accuracy \cite{berk2018fairness} & Examines whether the protected feature value affects the accuracy of the ML model's outcomes.\\
\hline
Equal opportunity \cite{hardt2016equality} & Given a positive ground truth label, examines whether the protected feature value affects the ML model's probability of outputting a positive ML model's outcome.\\
\hline
Treatment equality \cite{verma2018fairness} & Examines whether the protected feature value affects the ratio between the number of false negative (FN) and false positive (FP) ML model outcomes.\\
\hline
Equal false positive rate  \cite{berk2018fairness} & Examines whether the protected feature value affects the false positive rate (FPR) of the ML model's outcomes.\\
\hline
Equal false negative rate \cite{berk2018fairness} & Examines whether the protected feature value affects the false negative rate (FNR) of the ML model's outcomes.\\
\hline
Error rate balance \cite{narayanan2018translation} & Examines whether the protected feature value affects the FNR and FPR of the ML model's outcomes.\\
\hline
Normalized difference \cite{zliobaite2015relation} & Examines whether any of the protected feature values has an advantage in terms of obtaining a positive outcome from the ML model.\\
\hline
Mutual information \cite{fukuchi2015prediction} & Examines whether the protected feature contributes to the entropy of the ML model's outcomes, i.e., the effect of the protected feature on the ML model's outcome.\\
\hline
Balance residuals \cite{calders2013controlling} & Examines whether the protected feature value affects the error rate of the ML model's outcomes.\\
\hline
Calibration \cite{chouldechova2017fair} & Given a specific risk score value, examines whether the protected feature value affects the ML model's probability of outputting a positive ML model's outcome.\\
\hline
Prediction parity \cite{chouldechova2017fair} & Given a risk score value higher than a specific risk score threshold, examines whether the protected feature value affects the ML model's probability of outputting a positive ML model's outcome.\\
\hline
Error rate balance with score (ERBS) \cite{chouldechova2017fair} & Given a negative/positive ML model outcome, examines whether the risk score value is higher/lower than a specific risk score threshold.\\
\hline
\end{tabular}
\end{adjustbox}\caption{Existing bias detection and estimation methods and their ethical aspects}\smallskip\label{ethicsTable}
\end{table*}

In Table~\ref{ethicsTable}, we present the ethical aspects of the existing bias detection and estimation methods.

As noted, all bias detection and estimation methods are based on the ``fairness through unawareness" principle~\cite{verma2018fairness}, which means that changes in features with ethical significance should not change the ML model's outcome.
Each existing method examines only one ethical aspect derived from this principle, i.e., none of them perform a complete ethical evaluation.

In Table~\ref{ethicsTable}, we present the ethical aspect that is derived from the ``fairness through unawareness" principle for each method. 
Each method examines a selected statistical property according to the ML model's output, such as the ML model's true positive rate (TPR), true negative rate (TNR), outcome's entropy, etc.

As seen in Table~\ref{ethicsTable}, none of the existing methods evaluate the``fairness through unawareness" principle in its entirety.
In contrast, BENN performs a complete bias evaluation, as it examines the ML model for all ethical aspects derived from the ``fairness through unawareness" principle.

\section{Experimental Use Cases}
In our experiments we evaluated BENN on five different use cases:
\subsubsection{ProPublica COMPAS}~\cite{angwin2016machine}\footnote{github.com/propublica/compas-analysis/blob/master /compas-scores-two-years.csv} is a benchmark dataset that contains racial bias. 
The COMPAS system is used in the US by Florida's Department of Justice (judges and parole officers) to assess the likelihood of a defendant to be a repeat offender.
The dataset was collected from the COMPAS system's historical records of offenders from Broward County, Florida and contains the defendant's demographic characteristics, criminal history, and COMPAS recidivism risk scores (ranging between $[1,10]$, where one represents a low risk to commit another crime and 10 represents a high risk of doing so). 
ProPublica discovered that COMPAS tends to output low recidivism scores for Caucasian defendants, which results in discriminatory behavior toward any minority group in the dataset (such as African Americans or Asians).
The discrimination can also be observed by examining the prediction errors of COMPAS, as the rate of offenders that were labeled as high risk but did not commit another crime is $23.5\%$ for Caucasian defendants and $44.9\%$ for African American defendants~\cite{angwin2016machine}.
In addition, the rate of offenders that were labeled as low risk yet did commit another crime is $47.7\%$ for Caucasian defendants and $28.0\%$ for African American defendants~\cite{angwin2016machine}.
After filtering samples with missing values and nonmeaningful features, the dataset contains 7,215 samples and 10 features.

\subsubsection{Adult Census Income}~\cite{blake1998adult}\footnote{archive.ics.uci.edu/ml/datasets/adult} is a benchmark dataset that contains racial and gender-based bias. 
The Adult Census Income database contains census records that were collected between 1994 and 1995 by means of population surveys that were managed by the US Census Bureau.
The dataset corresponds to an income level prediction task, i.e., income above 50K or below.
After filtering samples with missing values and nonmeaningful features, the dataset contains 23,562 samples and 12 features.

\subsubsection{Statlog (German Credit Data)}~\cite{kamiran2009classifying}\footnote{archive.ics.uci.edu/ml/datasets/statlog+\\(german+credit+data)} is a benchmark dataset that contains gender-based bias.
The dataset contains customers' data, such as demographic information, account information,  out-of-bank possessions, and loan request information.
The dataset corresponds to the task of determining whether the customer should be considered for a loan.
After filtering samples with missing values and nonmeaningful features, the dataset contains 1,000 samples and 20 features.

\subsubsection{Telco Churn} Additional experiments were performed on a European Telco's churn prediction ML model and dataset.
This ML model is a DNN-based model that predicts customer churn, i.e., whether a customer will terminate his/her Telco subscription.
The model was trained and used internally, i.e., the model is not accessible through the Internet.
The Telco's churn dataset contains real customers' accounts and demographic information.
The data contains 95,704 samples and 28 features, and the protected feature is gender.
Bias presented in the churn prediction task might result in the company offering different incentives to different customers, depending on the customer's protected feature value.
For example, bias towards female customers (assuming they will not churn) could potentially result in offering better incentives to male customers.
In addition, bias in the churn prediction model may harm public relations and cause a reduction in revenue. 

\subsubsection{Synthetic Data} In order to provide a sanity check, we generated a synthetic dataset. 
The dataset contains three binary features, two of which are protected: one is a fair feature (has no bias) and one is extremely biased (has maximal bias), and a random feature. 
The dataset consists of 305 samples, which are composed from every possible combination of the feature values.
In order to construct the protected bias feature, we set each record value in the biased feature to be correlated with the true label, i.e., if the record label was one, then the bias feature value is set at one as well.
In order to construct the protected fair feature, we duplicate each possible record and set the fair feature at one for the first record and zero for the second.

\end{appendices}
\end{document}